\documentclass[letterpaper, 10 pt, conference]{ieeeconf}  

\IEEEoverridecommandlockouts
\usepackage[vlined, ruled, boxed, linesnumbered]{algorithm2e}

\SetCommentSty{mycommfont}

\SetKwProg{Function}{}{}{}

\usepackage{graphicx,subfig}
\usepackage{graphics}
\usepackage{amssymb,amsmath}
\usepackage{amssymb}
\usepackage[noend]{algpseudocode}
\usepackage[font={small}]{caption}

\usepackage[usenames, dvipsnames]{color}

\usepackage[normalem]{ulem}

\usepackage{tikz}
\usepackage{textcomp}
\usepackage{lipsum}
\usepackage[noadjust]{cite}
\usepackage{gensymb}

\usepackage{url}
\usepackage{hyperref}

\usepackage{booktabs}

\newcommand{\zbf}{\mathbf{z}}

\newcommand{\deltabf}{\boldsymbol{\delta}}

\title{\LARGE \bf \vspace{-2.5mm}
Overhead Image Factors for Underwater Sonar-based SLAM
\vspace{-2.5mm}

}

\author{John McConnell$^{1}$, Fanfei Chen$^{2}$ and Brendan Englot$^{1}$
\thanks{This research was supported in part by a grant from Schlumberger Technology Corporation, and in part by NSF grant IIS-1723996.}
\thanks{$^{1}$J. McConnell and B. Englot are with the Department of Mechanical Engineering, Stevens Institute of Technology, Hoboken, NJ, 07030, USA. {\tt\small$\{$jmcconn1,benglot$\}$@stevens.edu}}
\thanks{$^{2}$F. Chen is with Exyn Technologies, Philadelphia, PA, 19146, USA. {\tt\small frankycff@gmail.com}}
}

\begin{document}

\maketitle 
\thispagestyle{empty}
\pagestyle{empty}

\vspace{-2mm}

\begin{abstract} 
Simultaneous localization and mapping (SLAM) is a critical capability for any autonomous underwater vehicle (AUV). However,
robust, accurate state estimation is still a work in progress when using low-cost sensors. We propose enhancing a typical low-cost sensor package using widely available and often free prior information; overhead imagery. Given an AUV's sonar image and a partially overlapping, globally-referenced overhead image, we propose using a convolutional neural network (CNN) to generate a synthetic overhead image predicting the above-surface appearance of the sonar image contents. We then use this synthetic overhead image to register our observations to the provided global overhead image. Once registered, the transformation is introduced as a factor into a pose SLAM factor graph.
We use a state-of-the-art simulation environment to perform validation over a series of benchmark trajectories and quantitatively show the improved accuracy of robot state estimation using the proposed approach. We also show qualitative outcomes from a real AUV field deployment. Video attachment: \url{https://youtu.be/_uWljtp58ks}

\vspace{-2mm}

\end{abstract}

\section{Introduction} \vspace{-2mm}
Autonomous underwater vehicles (AUVs) provide critical capabilities for inspection, defense, and environmental monitoring \cite{SLB-2020}. However, unlike ground or aerial robotics, the perceptual sensors employed are often acoustic, providing robustness to variations in ambient lighting and water clarity in a relatively small package. For AUVs operating in cluttered environments, this leaves multi-beam sonar; profiling or imaging, as the acoustic perceptual sensor of choice. These sensors have an expansive field of view that can provide an AUV with panoramic situational awareness. However, they have a high cost relative to other robotics applications, as well as a low signal-to-noise ratio and low resolution. 

A critical capability for AUVs is state estimation. This often calls for the use of simultaneous localization and mapping (SLAM) to operate in unknown environments. However, many AUV state estimation systems rely on high cost, and highly accurate inertial navigation systems (INS). This means that for low-cost AUVs operating in challenging field settings, without high-grade INS, SLAM can be brittle, inaccurate, and will drift as the mission progresses. 

A potential solution for managing AUV state estimation error is active SLAM \cite{EM-2017}, \cite{Suresh-2020}, wherein a decision-making agent acknowledges the fragility of the state estimation framework and, when necessary, inserts waypoints or trajectories into its mission expected to curb the growth of uncertainty. Active SLAM often yields impressive results, but these come at a cost. So-called ``fly-through paths" where an AUV takes the most efficient path to the goal, performing SLAM en route, may require the insertion of detours to achieve loop closures. An alternative remedy would be to surface at regular intervals to receive GPS measurements. However, both potential solutions translate to less efficient AUV operations. Moreover, active SLAM and GPS are undesirable for tactical considerations \cite{GPS-jamming} when considering intelligence, surveillance, or reconnaissance (ISR) applications, when efficient transit to mission goals is a high priority. 

In this work we consider AUVs for which, per the aforementioned considerations, the vehicle needs to be free of trajectory constraints and operate without GPS. Moreover, a critical assumption in our work is that the surrounding environment contains structures that are commonly observable in both overhead and sonar images. We contend this is true for many applications in real-world settings, such as pier inspection, harbor patrol, ISR, or even windfarm navigation.

\begin{figure}[t]
\centering
\subfloat[Our experimental ``AUV": a custom-instrumented BlueROV2. 
\label{fig:leading_1}]{\includegraphics[height=3.8cm]{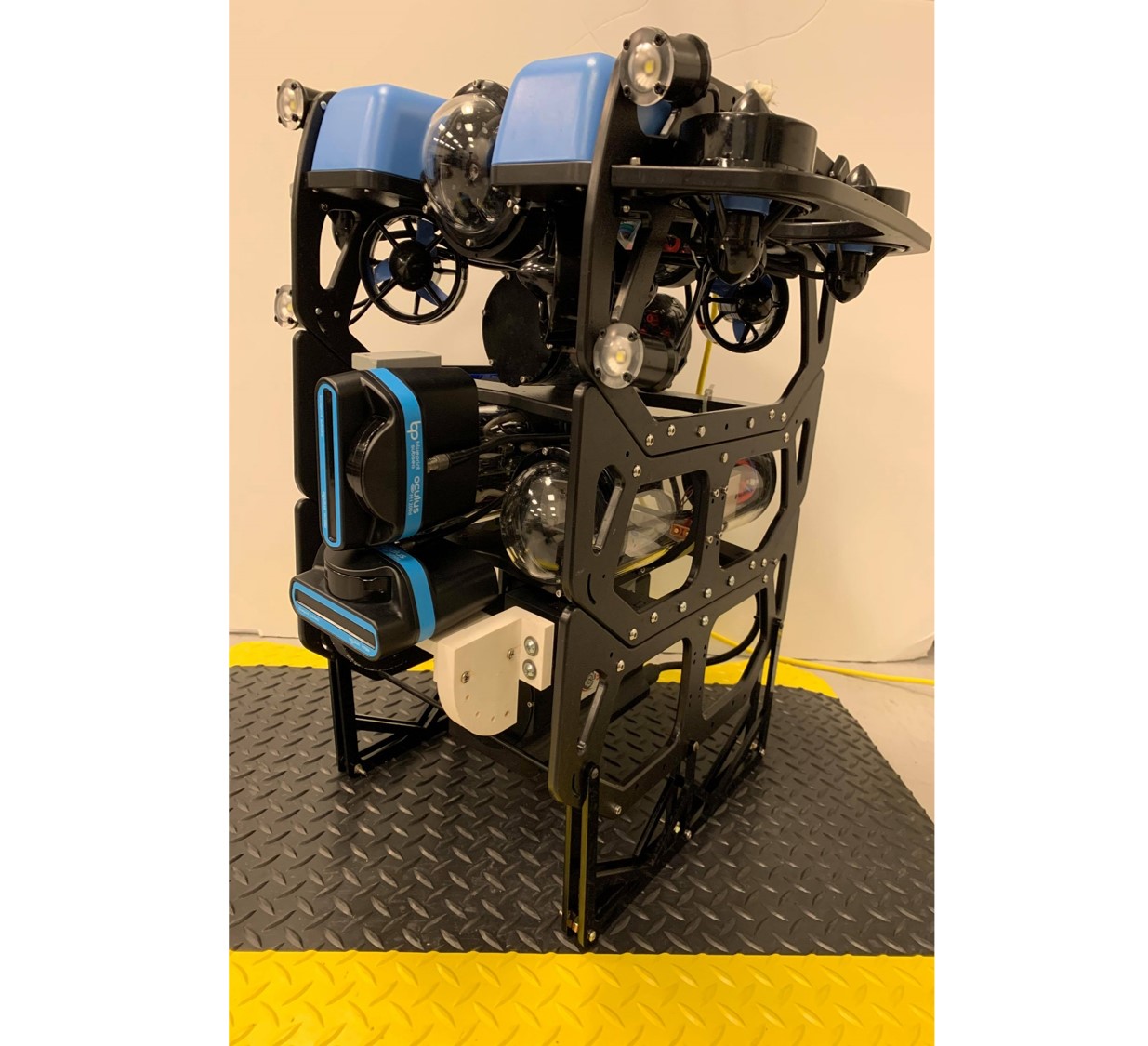}}\ \;
\subfloat[Simulated SLAM mission with OI factors. 
\label{fig:leading_2}]{\includegraphics[height=3.8cm]{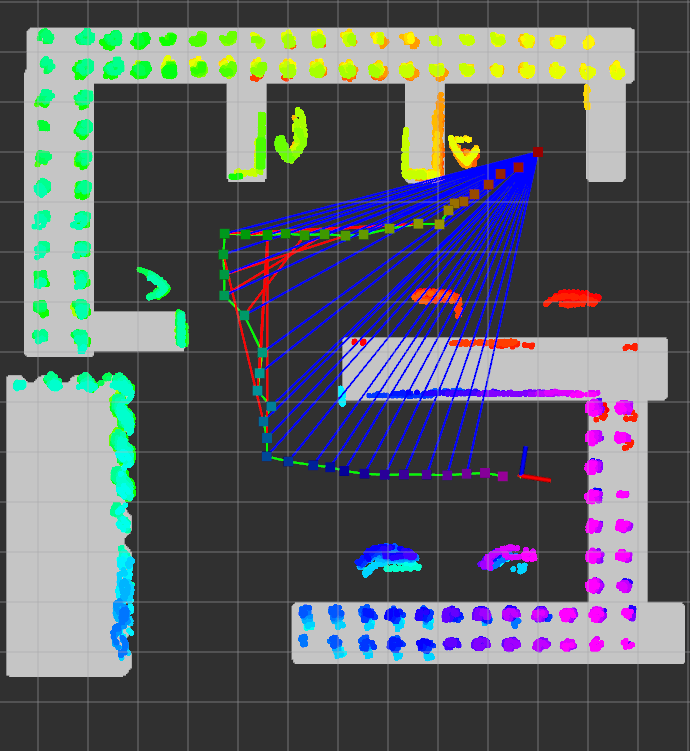}}\\ \vspace{-1mm}
\caption{\textbf{Highlights of the proposed method:} (a) the underwater robot used in our experiments; (b) SLAM with overhead image (OI) factors applied to harbor mapping in a high-fidelity simulation.}

\vspace{-6mm}
\label{fig:leading}
\end{figure}

Many methods have been proposed to improve underwater SLAM, however, none have applied overhead imagery to contribute measurement constraints to a SLAM solution. In this paper, we will examine the use of overhead 
imagery available for no cost in the public domain or from a low-cost unmanned aerial vehicle (UAV) as a method of improving AUV state estimation. Our contributions are as follows:
\begin{itemize}
    \item A novel representation and framework for the fusion of sonar and overhead RGB images. Unlike other work in this area, our method addresses the registration problem between overhead images and underwater sonar images.
    \item In contrast to other work in this area, we integrate the above into a pose SLAM framework incorporating odometry and sonar-based loop closures.
    \item We perform quantitative validation of the proposed method in a state of the art simulation environment, and give qualitative results from real robot deployments.
\end{itemize}

The paper structure is as follows: we will discuss related work, mathematically define the problem, present our proposed solution, and then give our results and conclusions.


\vspace{-1mm}
\section{Related Work}
\vspace{-1mm}

In subsequent sections we will define \textit{overhead image factors} and discuss their integration into an AUV's pose SLAM factor graph. First, we discuss relevant work that motivates both our choice of SLAM framework, and our decision to leverage overhead imagery for underwater SLAM.

\subsection{Underwater SLAM with Sonar} 

Graph-based pose SLAM has been used to support many successful underwater sonar-based SLAM applications. Ship hull inspection using a forward-looking imaging sonar is demonstrated in \cite{Li-2018}, multibeam profiling sonar SLAM traversing an underwater canyon is achieved in \cite{Hammond-2014}, and planar SLAM using imaging sonar observations of the seafloor is described in \cite{Johannsson-2010}. Additionally, \cite{Teixeira-2019} uses a factor graph to perform dense reconstruction of complex 3D structures using multibeam profiling sonar. Although underwater landmark-based SLAM with sonar has also been implemented successfully, \cite{Wang-2017,Westman-2018}, the challenge of landmark SLAM with acoustic sensors is the identification of point features, or the identification of objects as landmarks, and their subsequent, repeated association as they are re-observed.


In the work that follows, we adopt pose SLAM for this reason. There is no need to solve the data association problem over landmarks, and more importantly, this implies that landmark identification is not required, and sonar observations not amenable to conversion into landmarks (vessel hulls, seawalls, etc.) are still useful in the SLAM problem.

While all of the above case studies are successful, \cite{Li-2018}, \cite{Johannsson-2010} and \cite{Westman-2018} rely on a ring laser gyroscope for highly accurate, low-drift heading information. The inertial package in \cite{Teixeira-2019} also shows highly accurate results as it is a simulation derivative of the INS in \cite{Li-2018}, \cite{Johannsson-2010}, \cite{Westman-2018}. Lastly, \cite{Hammond-2014} and \cite{Wang-2017} use SLAM to address missing, or noisy inertial information, showing improvement, but with drift still present.  

While they represent impressive contributions to the state of the art, most of these works stand in contrast to our own, as they rely on expensive INS systems. In our work we focus on SLAM with a relatively low-cost vehicle, equipped with a Doppler velocity log (DVL) and a MEMS inertial measurement unit (IMU) for dead reckoning. 
While most SLAM frameworks (including ours) also take advantage of place recognition, a loop closure can only reduce drift to the level in the reference pose, not to zero. Therefore, our proposed use of overhead imagery is intended to address the fact that if loop closures are sparse, or an IMU is of a noisy low-cost nature, drift may grow to unacceptable levels.

\subsection{Fusion of Overhead Imagery and Ground Based Sensors}
The fusion of overhead imagery with ground-based sensors has been widely explored. Several paradigms have been applied, including the use of feature association and the use of CNNs.  Firstly \cite{Leung-2008}, \cite{Viswanathan-2014}, fuse overhead images with ground-level perspectives by finding likely image pairs using image descriptors. These probable image pairs are then leveraged in a particle filter to localize a robot. The authors of \cite{Workman-2015}, \cite{Kim-2017} and \cite{Shetty-2019} utilize convolutional neural networks (CNNs) to compute image similarity. A more apt comparison to our work is \cite{Tang-2020}, where live radar images are used to localize within a satellite image provided a priori, using deep learning to predict the rotation offset between images, and synthesize radar images to facilitate registration. In contrast, we use the output of overhead image registration to contribute factors to a SLAM factor graph, which is composed of a variety of measurement constraints from different sources. 


Overhead image fusion has also been explored in underwater robotics, fusing sonar imagery with overhead images using a CNN \cite{Machado-2020}. However, like  \cite{Workman-2015}, \cite{Kim-2017} and \cite{Shetty-2019}, image similarity is the learned output, and direct registration is not addressed, which is why we believe no comparison is warranted to \cite{Machado-2020}.
\cite{Giacomo-2021} proposes using a CNN to learn the mapping between a sonar image and the companion overhead RGB image. This idea performs well on the authors' similar data, but the authors 
note performance limitations when generalizing; 
we believe this is likely due to the poor correlation between
acoustic intensity and RGB values. Moreover, they do not consider the registration problem, of finding the transformation between a given satellite image footprint and the sonar image.
In contrast to these works, we will use a CNN to learn a more general image representation, separately address the registration problem, and use these measurements to improve an existing graph-based pose SLAM system that also incorporates other sources of measurement. 


\vspace{-1mm}

\section{Problem Description}
We formulate a three degree-of-freedom planar SLAM problem across discrete time steps $t$, each with an associated pose $\mathbf x_t$. We define each pose in the plane as 
\vspace{-1.5mm}
\begin{align}
    \mathbf x_t
    = \begin{pmatrix} x\ 
    y\
    \theta \end{pmatrix}^\top.
\end{align}
Note that we use bold lettering to define vectors and standard lettering to denote scalars. $\mathbb{R}$ denotes the set of real numbers and $\mathbb{R}_+$ denotes positive real numbers. Each pose has a set of observations  $\mathbf z_t$. The observations include sonar returns
in spherical coordinates with range $R\in \mathbb{R}_+$, bearing $\theta \subseteq [-\pi,\pi)$, and elevation $\phi \subseteq [-\pi,\pi)$ and an associated intensity value $\gamma \in \mathbb{R}_+$. These observations can be mapped into Cartesian space by 
\begin{align}
    \begin{pmatrix}  X \\  Y \\  Z \end{pmatrix} 
    = R\begin{pmatrix} \cos{\phi} \cos{\theta} \\ 
    \cos{\phi}\sin{\theta} \\ 
    \sin{\phi} \end{pmatrix}.
    \label{eq:to_cartesian}
\end{align}
Additionally, each set of observations $\zbf_t$ includes odometry information: ]linear velocity $V\in \mathbb{R}$, linear acceleration $A\in \mathbb{R}$ and rotational velocity $W\in \mathbb{R}$. In practice these measurements come from the DVL and IMU.

The robot moves through the environment at fixed depth, transitioning from state to state according to the dynamics
\vspace{-1.5mm}
\begin{align}
    \mathbf x_t = \mathbf g(\mathbf u_t , \mathbf x_{t-1}) + \deltabf_t,
\end{align}
where $\mathbf x_{t-1}$ is the pose at the previous timestep, $\mathbf u_{t}$ is the actuation command and $\deltabf_t$ is process noise. The posterior probability over the time history of poses is defined as 
\begin{align}
    p(\mathbf x_{1:t} , \mathbf m | \mathbf z_{1:t}, \mathbf u_{1:t}  ),
\end{align}
with map $\mathbf m$. The question addressed in this work, is how we can leverage the provided observations, in conjunction with reasonable prior information, to improve state estimation.  

\begin{figure}[t]
\centering
\subfloat[Simulated Sonar Image\label{fig:raw_sonar}]{\includegraphics[width=0.48\linewidth]{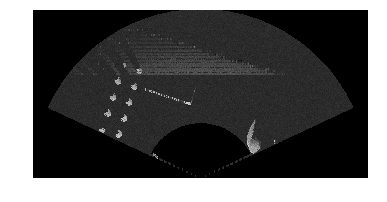}}\ \;
\subfloat[CFAR Image from (a)
\label{fig:CFAR}]{\includegraphics[width=0.48\linewidth]{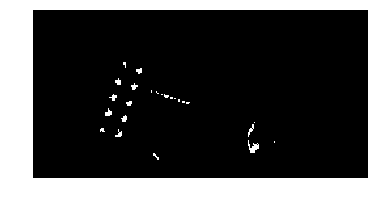}}\\
\subfloat[Candidate Overhead Image\label{fig:candiadte} ]{\includegraphics[width=0.48\linewidth]{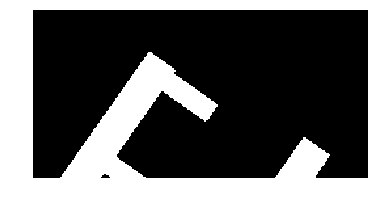}}\ \;
\subfloat[Example Predictions \label{fig:pred} ]{\includegraphics[width=0.48\linewidth]{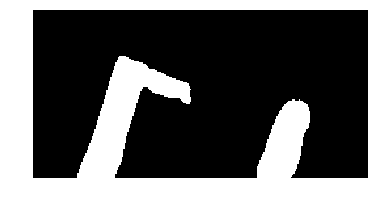}}
\caption{\textbf{Example Images:} (a) A sonar image from our simulation environment, (b) the CFAR-segmented sonar image, (c) the corresponding candidate overhead image, and
(d) the prediction of (b)'s appearance at the surface, from U-Net's processing of (b) and (c).}
\vspace{-6mm}
\label{fig:images}
\end{figure}


\section{Proposed Algorithm} 
\subsection{Sonar Image Processing}
As described in Section III, each sonar observation contains a set of  returns of varying intensity, in spherical coordinates, which are recorded in a 2D image. However, we must identify which points in the image are true sonar contacts and which are noise and/or second returns. 

In this work, we utilize constant false alarm rate (CFAR) detection \cite{Richards-2005} to identify returns from the surrounding environment in the sonar imagery. CFAR has been widely successful in radar and acoustic image processing \cite{El-Darymli-2018}, \cite{Acosta-2015}, and has supported our prior work in underwater active SLAM \cite{Wang-2021, Wang-2020}. Specifically, we use the smallest-of cell averaging (SOCA) variant of CFAR, which takes local area averages around the pixel in question and produces a noise estimate. If the signal is greater than a designated threshold, the pixel is identified as an image-feature. An example of its output is shown in Fig. \ref{fig:CFAR}; this perceptual data product is used to support SLAM throughout the work described in this paper.

All CFAR-extracted points in the image are converted to Cartesian coordinates (Eq. \eqref{eq:to_cartesian}). Because we confine our state to the plane, and because the elevation angle $\phi$ in Eq. \eqref{eq:to_cartesian} is not recorded in our sonar imagery, we set $\phi$ as zero. The consequence of the lack of elevation angle, is confinement of our system to in-plane pose estimation.

\begin{figure}[t]
\centering
{\includegraphics[width = .75\linewidth]{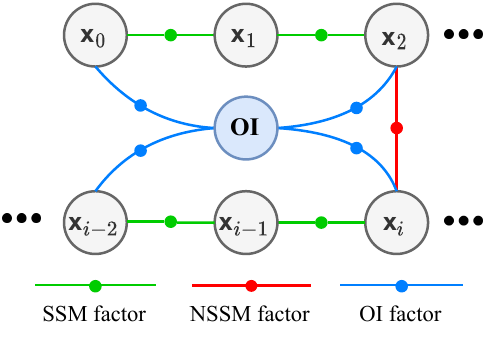}}
\caption{\textbf{SLAM Factor Graph.} Robot poses $x$ are connected by three kinds of factors: sequential scan matching factors (SSM) in green, non-sequential scan matching factors (NSSM, or loop closures) shown in red and overhead image factors (OI) shown in blue. 
}
\vspace{-6mm}
\label{fig:factor_graph}
\end{figure}

\vspace{-2mm}

\subsection{Graph Based Pose SLAM}
To estimate the location of the robot, we use the graph-based pose SLAM paradigm as in \cite{Wang-2021, Wang-2020}. At the front end, we utilize scan matching to introduce sonar-derived factors into the factor graph. We perform scan matching by using the iterative closest point (ICP) algorithm \cite{Besl-1992}. 
To initialize ICP with an accurate guess, the pose for a newly-arrived sonar frame is first predicted using our robot's DVL/IMU dead reckoning, and further optimized using consensus set maximization \cite{Fischler-1981}, helping ICP to avoid local minima.

At the back end, we utilize the GTSAM \cite{GTSAM} implementation of iSAM2 \cite{Kaess-2011}. Per Fig. \ref{fig:factor_graph}, we use three types of factors in our factor graph, sequential scan matching (SSM), non-sequential scan matching (loop closures - NSSM) and overhead image (OI) factors. The factor graph is denoted as 
\begin{flalign*}
\mathbf f(\boldsymbol \Theta) = \mathbf  f^{\text{0}}(\boldsymbol \Theta_0) & \prod_i \mathbf f^{\text{SSM}}_{i}(\boldsymbol \Theta_i) \prod_j \mathbf f^{\text{NSSM}}_j(\boldsymbol \Theta_j) \prod_q \mathbf f^{\text{OI}}_q(\boldsymbol \Theta_q).
\end{flalign*}
Sequential scan matching factors are produced by applying ICP to adjacent sonar keyframes, and non-sequential scan matching factors are developed by applying ICP between the current frame and frames inside a given search radius. Loop closure outliers are rejected by first evaluating point cloud overlap, and then applying pairwise consistent measurement set maximization (PCM) \cite{Mangelson-2018}. The proposed addition of overhead image factors will be discussed below. 

\subsection{Overhead Image Processing}
\begin{figure}[t]
\centering
{\includegraphics[width = .75\linewidth]{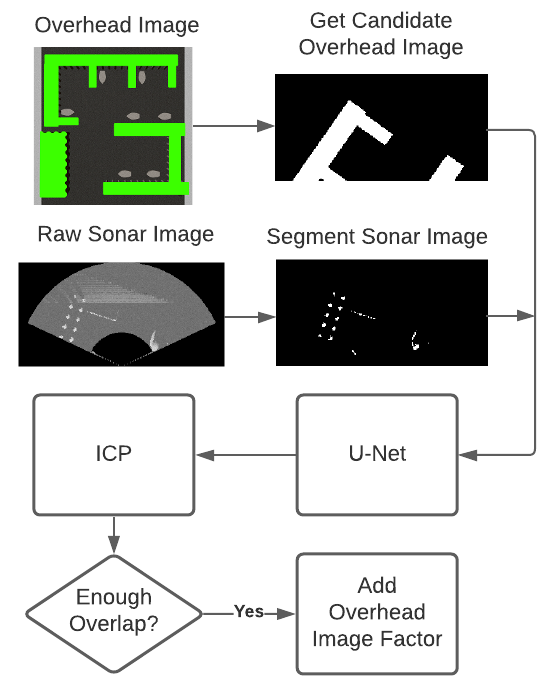}}
\caption{\textbf{System block diagram.} The overhead imagery is segmented offline, before the mission. As the mission progresses, a candidate overhead image is retrieved using the current state estimate. Raw sonar images are segmented using CFAR detection. Candidate overhead and segmented sonar images are presented in a query to U-Net. ICP is then applied, to register U-Net's synthetic image output to the globally referenced candidate overhead image. If there is enough overlap, the measurement is added to the factor graph.}
\vspace{-6mm}
\label{fig:flow}
\end{figure}

We segment the available overhead imagery into three classes: water, structures, and vessels. We then generate a binary mask of the overhead imagery; pixels comprising static structures are set to 1, and all others (water and vessels) are set to 0. An example of this mask is shown in green at the top left of Fig. \ref{fig:flow}. At each time-step, we carve out of the binary mask a footprint representing the sonar  field of view at the respective pose estimate, shown at top right of Fig. \ref{fig:flow}, and Fig. \ref{fig:candiadte}. This is performed using the current SLAM solution, in conjunction with the robot's known initial pose at the outset of the mission, to estimate the robot's current pose within the overhead imagery. The resulting image is referred to in this paper as the \textit{candidate overhead image}. 

\subsection{Learning a Unified Representation}
Key challenges to overcome in registering a sonar image to a candidate overhead image are the differences in perspective, contents, and modality. In general, an environment contains two kinds of structures: static structures, such as walls and docks, and dynamic structures, such as boats, which even when stationary, may be docked in different locations at different times. In this work, we assume dynamic structures may have moved between the time the overhead image was captured and the time of the mission, but not during the mission. Moreover, while overhead images contain RGB channels, sonar images only contain acoustic intensity, not necessarily related to the RGB channels of a given object. 

These fundamental differences are what makes registering a sonar image to an overhead RGB image challenging. For this reason, we propose to learn a unified representation that will aid in the registration problem. The first of our inputs is the candidate overhead image, with an example shown in Fig. \ref{fig:candiadte}. The second input is the sonar CFAR image, with an example shown in Fig. \ref{fig:CFAR}. These two images are stacked channel-wise into a single image as shown at left in Fig. \ref{fig:network}. Note that these images may have a non-zero transformation between them, as the state estimate may or may not have drifted. We would like to emphasize that their initial alignment does not need to be perfect, but simply provide reasonable overlap between the sonar image and candidate overhead image. The network output is the static structure observed in the overhead image transformed to the sonar image's frame; a synthetic overhead image. We utilize a well-known network architecture, U-Net \cite{Ronneberger-2015}, to learn the mapping between input and output with the addition of dropout layers. The key advantage of this method is learning a unified representation (a prediction of the above-surface appearance of the structures observed by the sonar) that can be registered to overhead imagery, offering the same perspective, contents and appearance as overhead imagery.

\vspace{-1mm}
\subsection{Acoustic Image Registration}
\vspace{-1mm}
We can now attempt to register the network output, a synthetic overhead image, to the candidate overhead image. To do so we first convert the images from pixels to meters. Next, we extract the outline of any structure present and apply voxel-based down-sampling to reduce the number of points. We then use ICP to estimate the transformation between the candidate overhead image points and the predicted synthetic overhead image points. 
The resulting transformation is used to estimate the overlap between the two sets of points; if the overlap does not exceed a designated threshold, it is considered to be a bad registration and discarded. Overlap is evaluated by applying nearest-neighbor association and counting the percentage of points in the source cloud with a Euclidean distance of less than one meter to their neighbor.  Note that here we do not apply PCM as a means of outlier rejection, as it is simply not required; the level of noise and ambiguity is not as severe as for standard loop closure detection. If the overlap is sufficient, then the ICP-derived overhead image transformation is added to the current SLAM state estimate to derive the transformation between the initial pose and the current pose. We then add this \textit{overhead image factor} to our factor graph, linking the initial and current pose. For the sake of generality, the factor graph in Fig. \ref{fig:factor_graph} shows an overhead image frame that can be associated with any point of reference, not just the initial robot pose.


\vspace{-1mm}
\section{Experiments and Results}
\vspace{-1mm}
\subsection{Hardware Overview and Simulation Environment}
\vspace{-1mm}
To perform real experiments, we utilize our custom-instrumented BlueROV2 heavy (Fig. 1a).  This vehicle is equipped with a pixhawk for stability control and a Jetson Nano for any required onboard computation. The vehicle is outfitted with a Rowe SeaPilot DVL, VectorNav VN100 IMU, and bar30 pressure sensor. We note that in contrast to many other works in this area, our vehicle is equipped with a low-cost MEMS IMU, rather than a high performance ring-laser gyroscope, which is our only source of heading information apart from the sonar and overhead imagery. For perception, we utilize the Blueprint Subsea Oculus M750d wide aperture multi-beam imaging sonar. This 750kHz sonar has a vertical aperture of 20\textdegree\, and a horizontal field of view of 130\textdegree. We operate the sonar at a max range of 30 meters. This relatively low-cost sensor stands in contrast to profiling sonars and higher-frequency imaging sonars more commonly used in this setting that offer higher resolution.
Our simulation environment reflects this sensor payload. We use Gazebo \cite{gazebo} in conjunction with sonar \cite{sonarsim} and UUV simulator \cite{uuvsim} packages to simulate an imaging sonar with the same characteristics as the M750d. 

\begin{figure*}[t]
\centering
{\includegraphics[height=4.6cm]{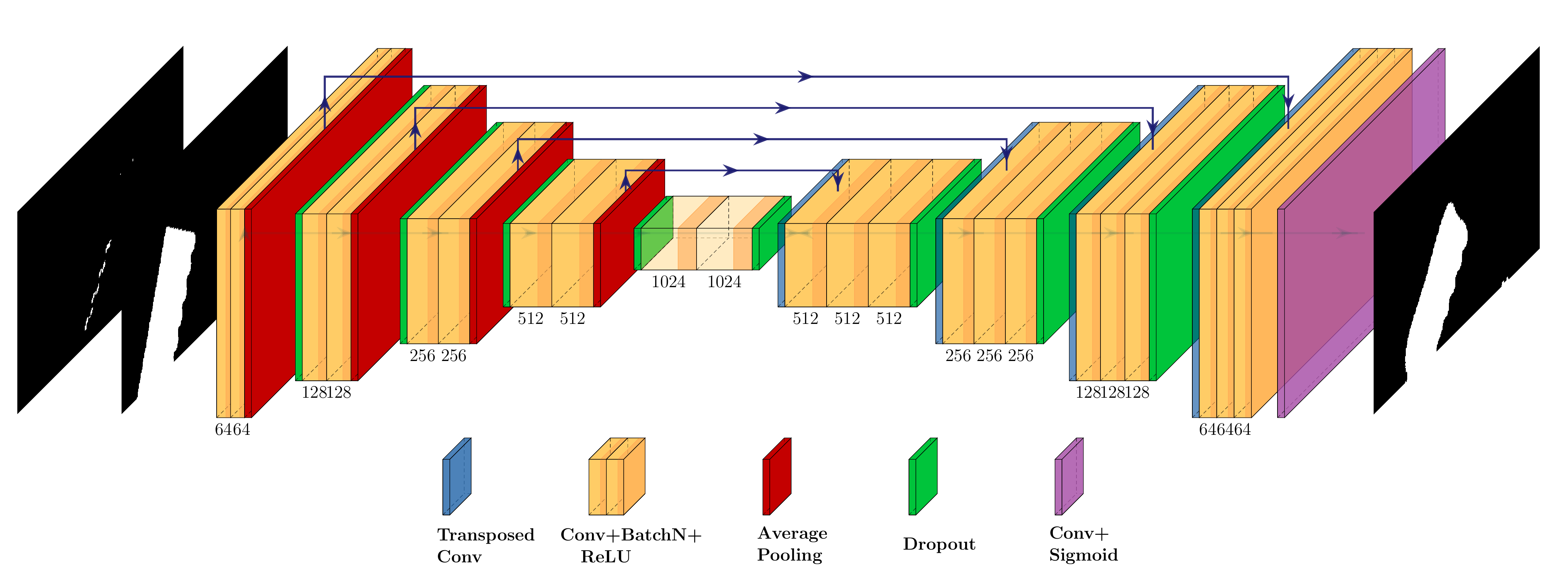}}
\caption{\textbf{Network Overview.} We utilize a U-Net architecture to generate synthetic overhead imagery predicting the above-surface appearance of an input sonar image's contents. The inputs to the network are a processed sonar image containing CFAR-detected features, and a candidate (partially overlapping) overhead image, each with the same spatial area. Input images are 256x128 pixels.}
\vspace{-6mm}
\label{fig:network}
\end{figure*}
\vspace{-1mm}

\subsection{Data Gathering}
To train our CNN, a significant amount of labeled data is required. While the authors of \cite{Machado-2020} and \cite{Giacomo-2021} use a sonar dataset with known locations from GPS, this data is not fully publicly available and is limited in scope, as it comes from a single environment with few dynamic structures. 
We perform a simulation study due to the lack of available data from underwater settings with ground truth, and our inability to perform new large-scale field experiments during the COVID-19 pandemic. Using a simulator allows us to generate a diverse body of relevant labeled imagery for our CNN, and to evaluate SLAM performance, as ground truth is known. Moreover, we will later demonstrate our algorithm's applicability to real-world data after being trained in simulation. Future work will focus on validating this methodology on real data from complex trajectories in large-scale environments with ground truth. 


To generate training data we use two environments, resembling a marina (Fig. \ref{fig:training_1}) and a ship's pier (Fig. \ref{fig:training_2}). To generate validation data, and to evaluate our SLAM solution, we design two other marina-like environments different from the first, shown in Figs. \ref{fig:testing} and \ref{fig:testing_2}. To gather data, we place the vehicle at a random, collision-free pose and capture a sonar image. We then generate a random transformation in-plane, with translations up to 5m 
and $\pm$ 22\textdegree\, yaw. We then capture a candidate overhead image at the random transformed pose. Labels are generated by capturing a candidate overhead image at the same pose as the sonar image. We generate 5000 samples from each training environment and 2500 samples from the validation environment in Fig. \ref{fig:testing} for ROC curve analysis. The environment in Fig. \ref{fig:testing_2} is used separately to evaluate long-duration SLAM performance.


Each training dataset requires a segmented overhead image of the full environment, for generating candidate overhead images as described in Sec. IV.C. Overhead images of each environment are generated at an altitude of 60m using a simulated camera. The aforementioned image segmentation (structure, water, vessels) is generated by hand labeling, as shown at the top left of Fig. \ref{fig:flow}. It may be desirable for this step in the process to be fully automated, but we also contend that with environments of this size, it may often be a reasonable cost to human AUV operators in the mission planning step, especially for settings that will be visited frequently. Moreover, deep learning methods for image segmentation in this setting are growing in maturity \cite{satellite}. 
Therefore, we confine the technical scope of this paper to include pre-segmented overhead imagery, 
as its preparation is simply a part of the mission planning process, 
and not an integral part of the live functionally of the proposed algorithm. 

\vspace{-1mm}

\subsection{CNN Training}
\begin{figure}[t]
\centering
{\includegraphics[width = 5.8cm]{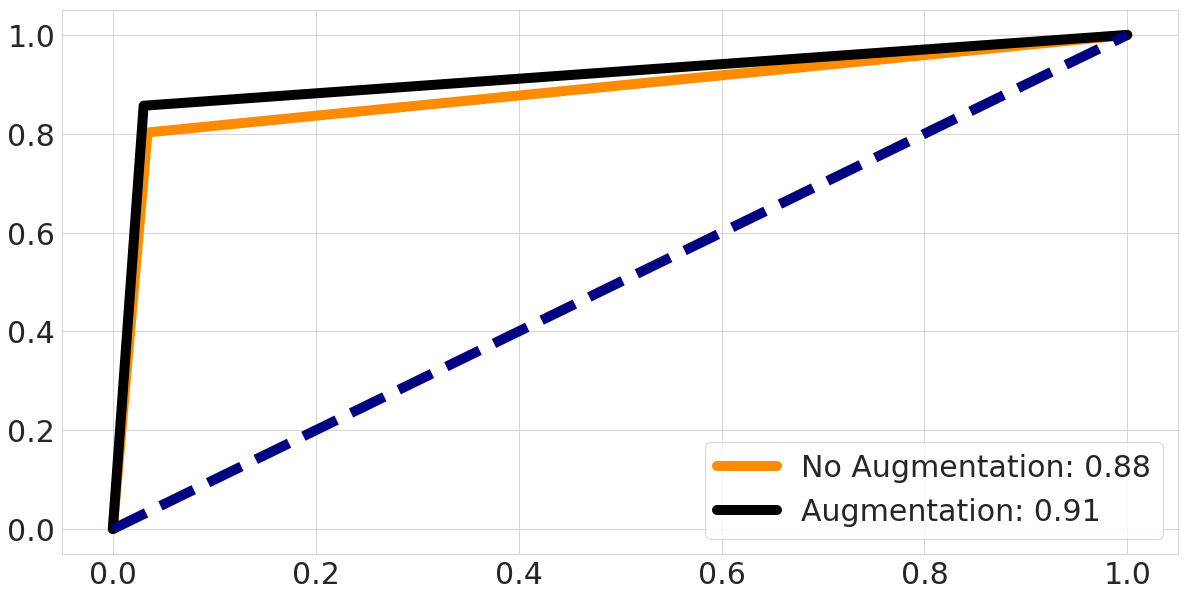}}
\caption{\textbf{Training ROC Curves.} Orange shows validation results without data augmentation; black shows validation results with data augmentation; the legend reports the area under the curve (AUC).} 
\vspace{-6mm}
\label{fig:roc}
\end{figure}

\begin{figure*}[t]
\centering
\subfloat[Example simulated SLAM mission with overhead image factors \label{fig:qual_proposed}]{\includegraphics[height=5.0cm]{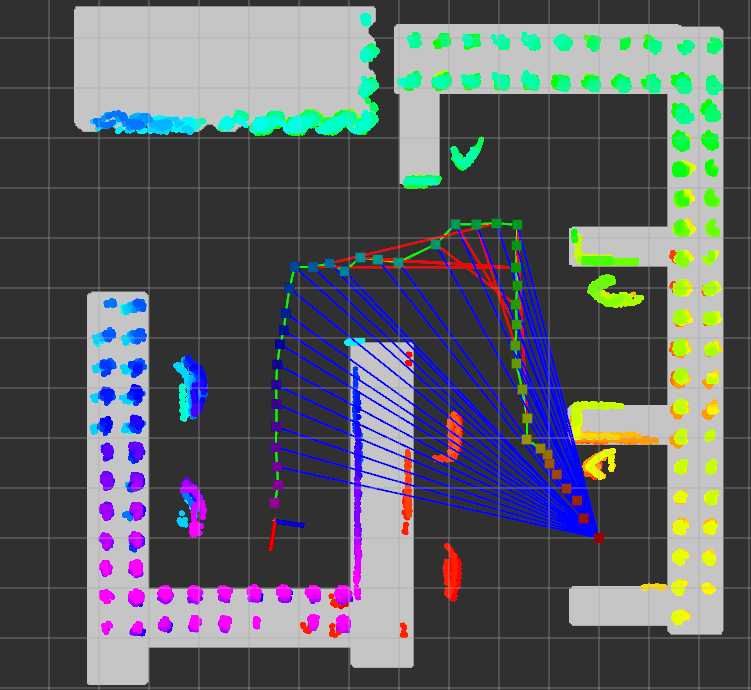}}\ \;
\subfloat[Example simulated SLAM mission without overhead image factors \label{fig:qual_baseline}]{\includegraphics[height=5.0cm]{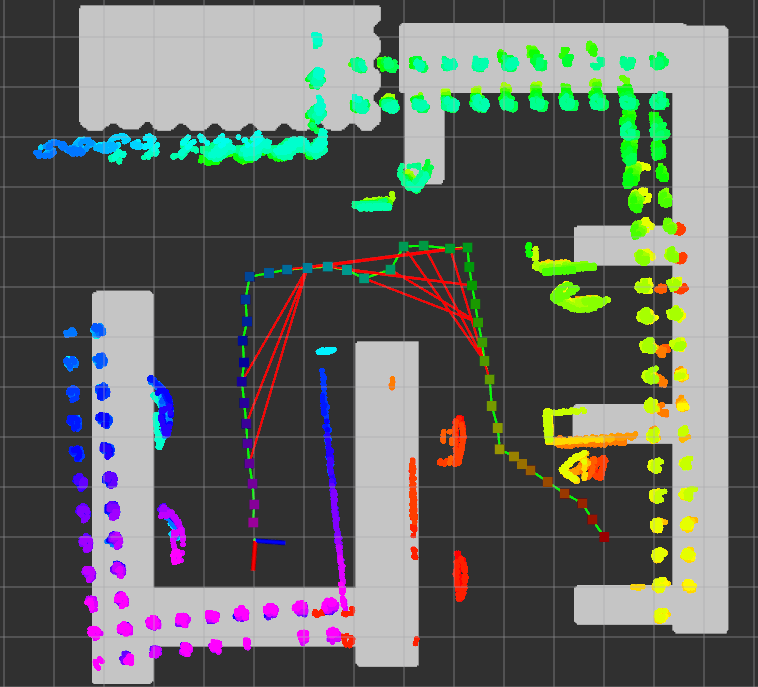}}\\
\subfloat[Example real-world SLAM mission with overhead image factors at SUNY Maritime \label{fig:qual_proposed_real}]{\includegraphics[width =.30\textwidth]{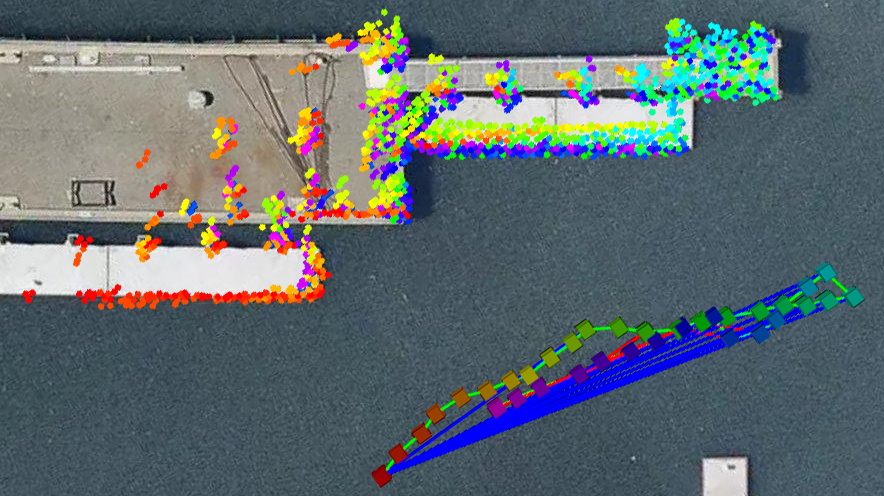}}\ \;
\subfloat[Example real-world SLAM mission without overhead image factors at SUNY Maritime \label{fig:qual_baseline_real}]{\includegraphics[width=.30\textwidth]{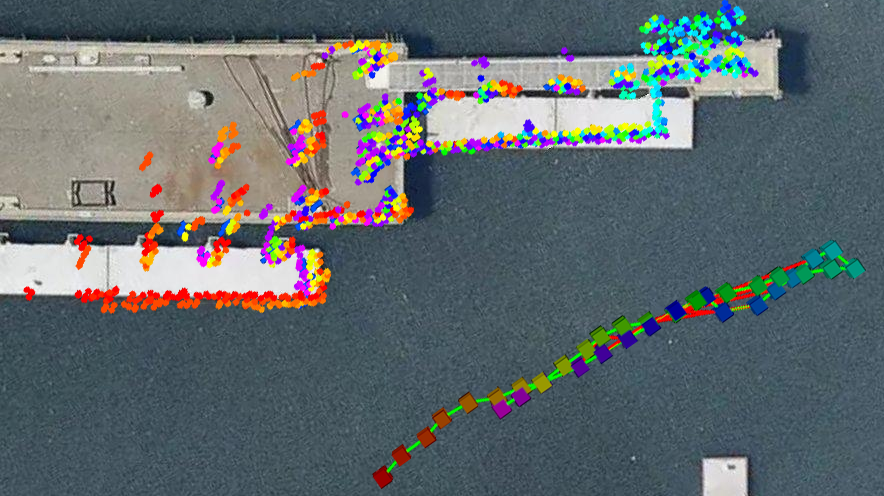}}\\
\subfloat[Example real-world SLAM mission with overhead image factors at USMMA \label{fig:qual_proposed_real_usmma}]{\includegraphics[width =.30\textwidth]{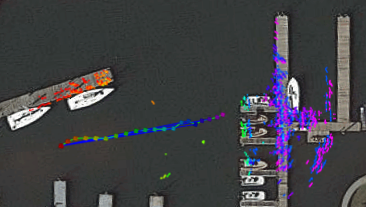}}\ \;
\subfloat[Example real-world SLAM mission without overhead image factors at USMMA \label{fig:qual_baseline_real_usmma}]{\includegraphics[width=.30\textwidth]{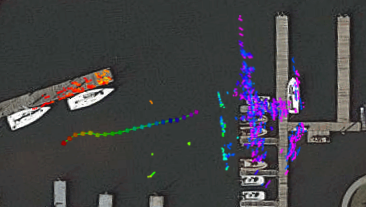}}\\
\subfloat[Training environment one \label{fig:training_1}]{\includegraphics[width=0.2\linewidth]{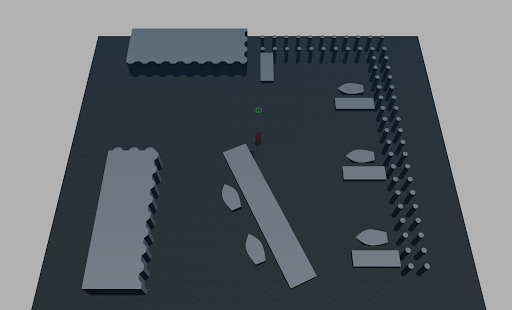}}\ \;
\subfloat[Training environment two \label{fig:training_2}]{\includegraphics[width=0.2\linewidth]{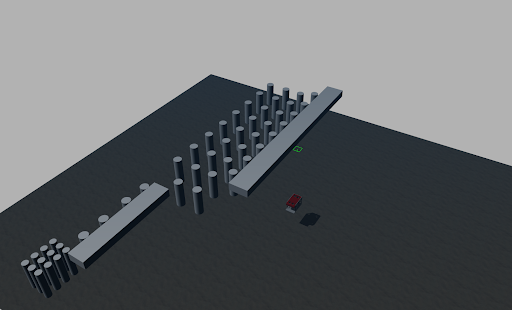}}\ \;
\subfloat[Validation environment one \label{fig:testing}]{\includegraphics[width=0.2\linewidth]{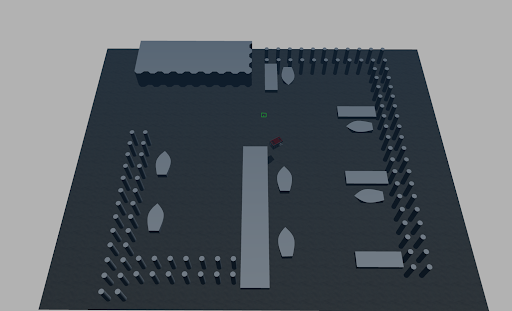}}\ \;
\subfloat[Validation environment two
\label{fig:testing_2}]{\includegraphics[width=0.2\linewidth]{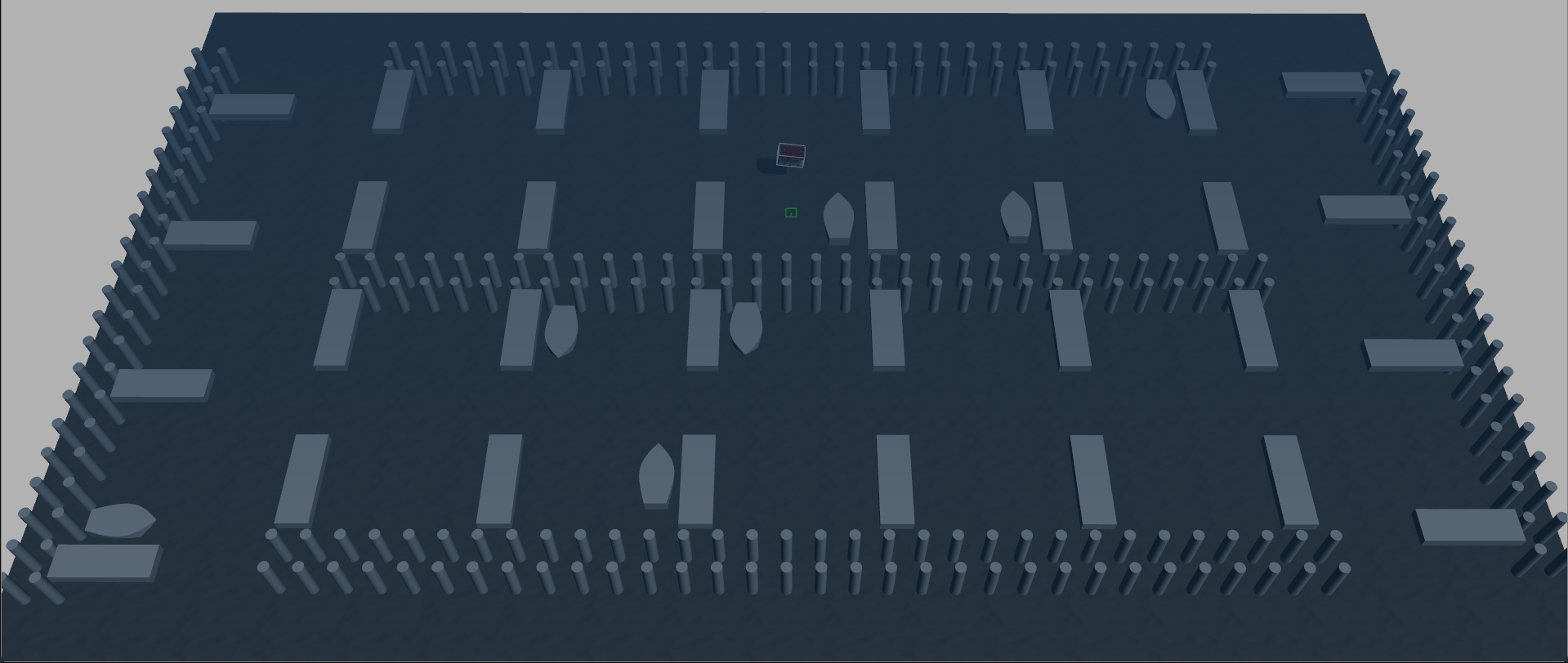}}\ \;
\vspace{-3mm}
\caption{\textbf{Example outcomes:} (a) SLAM mission with overhead image (OI) factors, and (b) the same trajectory without OI factors. The gray background of (a) and (b) shows the OI mask; grid boxes are 5m. (c) Results from the SUNY Maritime dataset using the proposed OI system, and (d) results without OI factors using the same trajectory shown in (c). (e) Results from USMMA with OI factors, and (f) results from the same trajectory without OI factors. SSM factors are shown as green lines, loop closures are shown as red lines, and OI factors are shown as blue lines. Poses are shown as  boxes, where color changes with time. The planar point cloud map utilizes the same color scheme as the poses the points are derived from. (g),(h) Our simulation training environments, and (i),(j) our validation environments. } 
\vspace{-5mm}
\label{fig:qual_results}
\end{figure*}

Using the gathered dataset, we implement the model architecture illustrated in Fig. \ref{fig:network} using TensorFlow \cite{tensorflow}. The model is trained for 10 epochs using a categorical cross-entropy loss function and the Adam optimizer \cite{Kingma-2015}. We evaluate our model using the validation dataset. 
We note that our validation environment (pictured in Fig. \ref{fig:testing}) 
has a difference in static structure, and its vessels 
are all in new locations, relative to the training environments. To quantitatively measure model performance we use receiver operator characteristic (ROC) curves, shown in Fig. \ref{fig:roc}.
To increase validation performance, we perform two types of simple data augmentation. First, we randomly flip each set of images. Secondly, we introduce Gaussian speckle noise to the set of input images. More advanced data augmentation (rotations, etc.) is not used to preserve the contents of each data point, as our imagery is often sparse on the sonar side. 

\vspace{-2mm}

\subsection{SLAM Performance Metrics} \vspace{-1mm}
To evaluate the performance of a SLAM framework aided by overhead image factors, we consider two key metrics, mean absolute error (MAE) and root mean squared error (RMSE). These metrics are chosen to not only characterize the mean of the error distributions but the goodness of fit the SLAM solution provides with respect to the true trajectory. We consider both the Euclidean distance of our SLAM estimate from the corresponding true pose, and the difference in estimated yaw angle $\theta$ from that of the true pose. Note that we do not consider the global pose, but the pose relative to the initial keyframe, ensuring a fair comparison across competing algorithms. In the results to follow, sonar imagery is sampled at 5Hz, and as in \cite{Wang-2021, Wang-2020}, sonar keyframes are generated upon every 2m of translation or 30\textdegree\, of rotation.

\vspace{-2mm}

\subsection{Simulated SLAM Results} \vspace{-1mm}
In addition to sonar imagery, our SLAM method requires a source of dead reckoning, in this case, a simulated DVL and IMU. Zero mean Gaussian noise is introduced into these sensors with $\sigma_{velocity}$ and $\sigma_{theta}$ as 0.1 m/s and 1\textdegree\, respectively. These values are chosen as they produce results similar to those experienced in the field with our baseline method. Lastly, the proposed method requires a robot position fix defined within the overhead imagery. We denote this as the overhead image frame, which we choose to coincide with the location of the robot's initial pose, defined with respect to the overhead imagery. Each mission is initialized by applying zero-mean Gaussian noise to the true initial pose using $\sigma_{theta}$, and $\sigma_{xy}$ of 1 meter. Note that in a real world mission, we believe this information could be easily sourced by collecting an initial GPS fix at the surface, at the beginning of a mission. Recall that we use point-cloud overlap as a means of outlier rejection, and for this simulation study, the minimum required overlap to accept an overhead image factor is 80\%.  

Equipped with a trained CNN, we can apply our full proposed pipeline to the validation environment of Fig. \ref{fig:testing}. To evaluate our SLAM system we generate two ``fly-through" trajectories at a fixed depth of approximately 1 meter; 
one transits from left to right and the other from right to left. This style of trajectory is selected since it efficiently transits the environment, while observing nearly all structures in the marina along the way. Using these two trajectories, we evaluate our proposed SLAM method as well as a baseline method, which is simply our SLAM method without the addition of overhead image factors. 
The performance of each competing method is evaluated on both fly-through trajectories, and a summary of results is given in Table \ref{table:results}. When considering results from the fly-through trajectories, even with numerous loop closures, the SLAM solution without overhead image factors still accumulates error as shown in Fig. \ref{fig:qual_baseline}. Conversely, the addition of the overhead image factors in the fly-through trajectory case reduces all error metrics and produces a more accurate 
map, shown in Fig. \ref{fig:qual_proposed}. 

Secondly, in order to evaluate the efficacy of our algorithm over long distances, we run a 2.7-kilometer ``long distance'' trajectory using the simulation environment shown in Fig. \ref{fig:testing_2}. This trajectory is generated by looping around the environment ten times.
With overhead image factors, a three-fold improvement is shown in terms of position, with a similar yaw MAE and marginal increase in yaw RSME. Results are summarized in Table \ref{table:results}. We note that no modifications are made to our implementation when running the long-distance trajectory experiment, and online performance is still achieved.

\begin{table}[h]
\vspace{-0mm}
\centering
\begin{tabular}{ccccc}
\toprule
& \multicolumn{2}{c}{Fly Through} & \multicolumn{2}{c}{Long Distance}\\
Metric & Proposed & Baseline & Proposed & Baseline\\
\midrule
Euclidean Error&&&&\\
MAE (meters) & \bf 0.83 & 2.06 & \bf0.87 & 3.13\\
RMSE (meters) & \bf 0.95 &  2.29 & \bf0.93 & 3.17 \\
\midrule
Yaw Angle Error&&&&\\
MAE (degrees) & \bf 1.70 & 5.95 & \bf 3.65 & 3.98\\
RMSE (degrees) & \bf 2.09  &  6.48 & 10.5 & \bf 10.27 \\
\toprule
\end{tabular}
\caption{\textbf{SLAM Results.} MAE and RMSE. ``Fly through" refers to the two benchmark trajectories in validation environment one as in Fig. \ref{fig:testing}. ``Long distance" refers to the 2.7km transit through validation environment two shown in Fig. \ref{fig:testing_2}. }
\vspace{-6mm}
\label{table:results}
\end{table}

\subsection{Real World SLAM Results}

While we confine our quantitative case study to simulation, we demonstrate the promise of future work by testing our system on real data. Due to restrictions resulting from the COVID-19 pandemic, gathering a new large-scale, ground-truthed dataset for this study was not possible; instead, previously-gathered datasets were utilized.
These feature data collected with our custom BlueROV2 in two marina environments: the SUNY Maritime College marina on the East River in The Bronx, NY and the U.S. Merchant Marine Academy (USMMA) on the Long Island Sound in Kings Point, NY. Our training maps, shown in Figs. \ref{fig:training_1} and \ref{fig:training_2}, are intended to capture the general appearance of these environments. These real-world environments pose several challenges to our overhead image factor system. Firstly, we train in simulation, and it is unknown how well our trained CNN will generalize to the real world. Secondly, the simulator, while realistic, is a controlled environment. Moreover, our real-world settings present environmental challenges, such as currents up to 2kts.

The datasets are recorded at a fixed depth of approximately 2m for SUNY Maritime and 1m for USMMA. Overhead imagery is hand segmented and provided to the system a priori, in addition to the robot's initial location in the imagery. The initial location was determined by manual alignment. Again we use point-cloud overlap as a means of outlier rejection, and for these field datasets, the minimum required overlap to accept an overhead image factor is 95\%.  

At SUNY Maritime, we consider an ``out and back'' trajectory where the robot moves away from its starting location and returns along a similar path within close proximity of the starting location. Baseline results for SUNY Maritime are shown in Fig. \ref{fig:qual_baseline_real}. Although our system was trained in simulation, it contributes several overhead image factors, mitigating the drift compared to the baseline method as shown in Fig. \ref{fig:qual_proposed_real}. The baseline method mainly shows drift in $\theta$, and even though several loop closures occur, error still accumulates in the map.

At USMMA, we consider a different trajectory, a straight-line transit where no loop closures occur in the standard SLAM solution. We consider this type of trajectory to demonstrate the utility of overhead image factors, which can correct drift even when loop closures are unavailable. 
The drift of the baseline trajectory, shown in Fig. \ref{fig:qual_baseline_real_usmma}, results in a highly inaccurate map. Once again our system, trained only in simulation, can contribute several overhead image factors that improve the state estimate, as shown in Fig. \ref{fig:qual_proposed_real_usmma}. 

We wish to underscore the significance of these results and the \textit{potential} of the proposed framework. Although the system was trained in simulation, it can, and in these two environments, does generalize to real sonar data. However, we are keenly aware of the challenges associated with deploying deep learning methods in the field. Future work will focus on expanding our methodology and testing on more significant, more complex trajectories and developing a ground-truthed dataset. Full video playback of these experiments, in addition to portions of our simulated SLAM experiments, are included in our \textcolor{blue}{\href{https://youtu.be/_uWljtp58ks}{video attachment}}.

\subsection{Computation Time}
Our framework results in the following sources of additional computational overhead: a single GPU query, a single ICP query, and a single instance of overlap estimation per keyframe, which are implemented in a multi-threaded manner. 
If an overhead image factor is detected, the result is passed to the SLAM thread, where it is added to the factor graph. The result is a SLAM system that runs faster than keyframes are added. A runtime test of our overhead image factor system, which encompassed network prediction, ICP, and overlap estimation, yielded a rate of 55Hz, which is above and beyond the requirements of sonar keyframe processing. 
This test was performed over the 2500 examples from our validation set. These experiments were conducted using an Intel i9-9900K CPU @ 3.60GHz and an Nvidia Titan RTX GPU. 

\vspace{-4mm}

\section{Conclusions}
In this work, we have presented a novel method for addressing drift in a sonar-based underwater SLAM solution, using overhead image factors. We have shown that our method provides significant added value in terms of pose estimation error. Moreover, we do not use complex prior information such as environment CAD models. Instead we use data that can be sourced from the public domain, and an initial position fix. Additionally, we do not attempt to localize a robot using only an overhead image; we use all of the available information in conjunction with state-of-the-art sensor fusion methods \cite{GTSAM} to build a robust and low drift state estimate. We do concede that this method is mostly applicable in littoral environments, where structures are observable above and below the water. However, we would contend that this represents many AUV/ROV applications, such as hull inspection, harbor security, and operating near offshore structures. When considering potential shortfalls of this system, we make one key assumption throughout this work; that the surrounding environment contains structures observable by both sonar and overhead imagery. Otherwise, our method will be unable to mitigate drift using overhead image factors. Moreover, the same will be true if the structures in question are degenerate 
(consider a single small pier piling or a long flat wall).
Lastly, in this work, we train and evaluate primarily in simulation. Future work will focus on developing the data (simulation and real) to extend this work to more extensive, outdoor, real world environments in the littoral zone and near offshore energy assets.


\vspace{-2mm}

{}


\begin{thebibliography}{99}

\bibitem{SLB-2020}
J. Vincent \textit{et al.},
``Supervised Multi-Agent Autonomy for Cost-Effective Subsea Operations,'' \textit{
Proc. Offshore Technology Conf.}, 2020. 

\bibitem{EM-2017} J. Wang  and  B. Englot,  ``Autonomous  exploration  with  expectation-maximization,''   \textit{Proc. Int. Symp. Robotics Res.}, 2017.

\bibitem{Suresh-2020} S. Suresh, P. Sodhi, J. Mangelson, D. Wetttergreen, and  M. Kaess, ``Active  SLAM  using  3D  submap  saliency  for  underwater  volumetric exploration,''  \textit{Proc. IEEE Int. Conf. Robotics Automation}, 3132-3138, 2020.

\bibitem{GPS-jamming} A. Grant, P. Williams, N. Ward and S. Basker, ``GPS Jamming and the Impact on Maritime Navigation,'' \textit{J. Navigation}, 62(2), 173-187, 2009.

\bibitem{Li-2018}
J. Li, M. Kaess, R.M. Eustice and M. Johnson-Roberson, ``Pose-Graph SLAM Using Forward-Looking Sonar," \textit{IEEE Robotics and Automation Lett.}, 3(3), 2330-2337, 2018.

\bibitem{Hammond-2014}
M. Hammond and S. M. Rock, ``A SLAM-based approach for underwater mapping using AUVs with poor inertial information," \textit{Proc. IEEE/OES Autonomous Underwater Vehicles Symp.}, 2014.

\bibitem{Johannsson-2010}
H. Johannsson, M. Kaess, B. Englot, F. Hover and J. Leonard, ``Imaging sonar-aided navigation for autonomous underwater harbor surveillance," \textit{Proc. IEEE/RSJ Int. Conf. Intelligent Robots Syst.}, 4396-4403, 2010.

\bibitem{Teixeira-2019}
P. V. Teixeira, D. Fourie, M. Kaess and J.J. Leonard, ``Dense, Sonar-based Reconstruction of Underwater Scenes," \textit{Proc. IEEE/RSJ Int. Conf. Intelligent Robots Syst.}, 8060-8066, 2019.

\bibitem{Wang-2017}
J. Wang, S. Bai and B. Englot, ``Underwater localization and 3D mapping of submerged structures with a single-beam scanning sonar," \textit{Proc. IEEE Int. Conf. Robotics Automation}, 4898-4905, 2017.

\bibitem{Westman-2018}
E. Westman, A. Hinduja and M. Kaess, ``Feature-Based SLAM for Imaging Sonar with Under-Constrained Landmarks," \textit{Proc. IEEE Int. Conf. Robotics Automation}, 3629-3636, 2018.

\bibitem{Leung-2008}
K.Y.K. Leung, C.M. Clark and J.P. Huissoon, ``Localization in urban environments by matching ground level video images with an aerial image," \textit{Proc. IEEE Int. Conf. Robotics Automation}, 551-556, 2008.

\bibitem{Viswanathan-2014}
A. Viswanathan, B.R. Pires and D. Huber, ``Vision based robot localization by ground to satellite matching in GPS-denied situations," \textit{Proc. IEEE/RSJ Int. Conf. Intelligent Robots Syst.}, 192-198, 2014.

\bibitem{Workman-2015}
S. Workman, R. Souvenir and N. Jacobs, ``Wide-Area Image Geolocalization with Aerial Reference Imagery," \textit{Proc. IEEE Int. Conf. Computer Vision}, 3961-3969, 2015.

\bibitem{Kim-2017}
D. Kim and M.R. Walter, ``Satellite image-based localization via learned embeddings," \textit{Proc. IEEE Int. Conf. Robotics Automation}, 2073-2080, 2017.

\bibitem{Shetty-2019}
A. Shetty and G.X. Gao, ``UAV Pose Estimation using Cross-view Geolocalization with Satellite Imagery," \textit{Proc. IEEE Int. Conf. Robotics Automation}, 1827-1833, 2019.

\bibitem{Tang-2020}
T.Y. Tang, D. De Martini, D. Barnes and P. Newman, ``RSL-Net: Localising in Satellite Images From a Radar on the Ground," \textit{IEEE Robotics and Automation Lett.}, 5(2), 1087-1094, 2020.

\bibitem{Machado-2020}
M. Machado Dos Santos, G.G. De Giacomo, P.L.J. Drews and S.S.C. Botelho, ``Matching Color Aerial Images and Underwater Sonar Images Using Deep Learning for Underwater Localization," \textit{IEEE Robotics and Automation Lett.}, 5(4), 6365-6370, 2020.

\bibitem{Giacomo-2021}
G.G. De Giacomo, M.M. dos Santos, P.L.J. Drews and S.S.C. Botelho, ``Guided Sonar-to-Satellite Translation," \textit{J. Intelligent and Robotic Syst.}, 101, 46, 2021. 

\bibitem{Richards-2005}
M. Richards, \textit{Fundamentals of Radar Signal Processing}, 2005.

\bibitem{El-Darymli-2018}
K. El-Darymli, P. McGuire, D. Power and C. Moloney, ``Target detection in synthetic aperture radar imagery: a state-of-the-art survey,'' \textit{J. Applied Remote Sensing}, 7(1), 6014-6058, 2013.

\bibitem{Acosta-2015}
G. Acosta and S. Villar, ``Accumulated CA–CFAR Process in 2-D for Online Object Detection From Sidescan Sonar Data,'' \textit{IEEE J. Oceanic Engineering}, 40(3), 558-569, 2015.

\bibitem{Wang-2021}
J. Wang, F. Chen, Y.Huang, J. McConnell, T. Shan, B. Englot, “Virtual Maps for Autonomous Exploration of Cluttered Underwater Environments,”
\textit{IEEE Int. Conf. Robotics Automation (ICRA) Workshop on Underwater Active Perception}, 2021.

\bibitem{Wang-2020}
J. Wang, \textit{Towards 3D Mapping and Autonomous Exploration for Underwater Robots in Cluttered Environments}, Ph.D. Thesis, Stevens Institute of Technology, 2020. 

\bibitem{Besl-1992}
P.J. Besl and N.D. McKay, “A Method for Registration of 3-D Shapes,”
\textit{IEEE Trans. Pattern Analysis and Machine Intelligence},
14(2), 239–256, 1992.

\bibitem{Fischler-1981}
M. Fischler and R. Bolles, “Random Sample Consensus: A Paradigm
for Model Fitting with Application to Image Analysis and Automated
Cartography,” \textit{Communications of the ACM}, 24(6), 381-395, 1981.

\bibitem{GTSAM}
F. Dellaert, ``Factor Graphs and GTSAM A Hands-on Introduction," Georgia Institute of Technology, Tech. Rep. No. GT-RIM-CPR-2012-002, 2012. 

\bibitem{Kaess-2011}
M. Kaess, H. Johannsson, R. Roberts, V. Ila, J. Leonard, and F. Dellaert, ``iSAM2: Incremental Smoothing and Mapping using the Bayes Tree," \textit{Int. J. Robotics Res.}, 31(2), 216-235, 2012.

\bibitem{Mangelson-2018}
J.G. Mangelson, D. Dominic, R.M. Eustice and R. Vasudevan, ``Pairwise Consistent Measurement Set Maximization for Robust Multi-Robot Map Merging," \textit{Proc. IEEE Int. Conf. Robotics Automation}, 2916-2923, 2018.

\bibitem{Ronneberger-2015}
O. Ronneberger, P.Fischer and T. Brox, "U-Net: Convolutional Networks for Biomedical Image Segmentation", \textit{Proc. Int. Conf. Medical Image Computing Computer-Assisted Intervention}, 234-241, 2015.

\bibitem{gazebo}
N. Koenig and A. Howard, ``Design and Use Paradigms for Gazebo, an Open-source Multi-robot Simulator," \textit{Proc. IEEE/RSJ Int. Conf. Intelligent Robots Syst.}, 3, 2149-2154, 2004.

\bibitem{sonarsim}
R. Cerqueira, T. Trocoli, G. Neves, S. Joyeux, J. Albiez, and L. Oliveira, ``A Novel GPU-based Sonar Simulator for Real-time Applications," \textit{Computers and Graphics}, 68, 66-76, 2017. 

\bibitem{uuvsim}
M.M.M. Manhães, S.A. Scherer, M. Voss, L.R. Douat and T. Rauschenbach, ``UUV Simulator: A Gazebo-based Package for Underwater Intervention and Multi-robot Simulation," \textit{Proc. IEEE/MTS OCEANS Conf.}, 2016.

\bibitem{satellite}
I. Demir \textit{et al.}, 
``DeepGlobe 2018: A Challenge to Parse the Earth Through Satellite Images," \textit{Proc. IEEE Conf. Computer Vision Pattern Recognition (CVPR) Workshops}, 172-181, 2018.


\bibitem{tensorflow}
M. Abadi \textit{et al.}, ``Tensorflow: Large-scale machine learning on heterogeneous distributed systems," \textit{arXiv preprint arXiv:1603.04467}, 2016.

\bibitem{Kingma-2015}
D.P. Kingma and J. Ba, ``Adam: A Method for Stochastic Optimization," \textit{Proc. Int. Conf. Learning Representations}, 2015.



\end{thebibliography}
\end{document}